\documentclass[letterpaper, 10 pt, conference]{ieeeconf}  % Comment this line out if you need a4paper

\IEEEoverridecommandlockouts                              % This command is only needed if 
\usepackage{hyperref}
\usepackage{url}
\usepackage{cite}
\usepackage{subfigure}
\usepackage{graphicx}
\usepackage{amsfonts}
\usepackage{booktabs} 
\usepackage{algorithm}
\usepackage{algpseudocode}
\usepackage[dvipsnames]{xcolor}
\usepackage{glossaries}

\newcommand{\cmt}[1]{}

\long\def\ignorethis#1{}

\newcommand{\etal}{{\em{et~al.}\ }}
\newcommand{\eg}{e.g.\ }

\newcommand{\vc}[1]{\ensuremath{\mathbf{#1}}}

\newcommand{\pctab}{\hspace{0.2in}}

\makeglossaries

\newglossaryentry{observation}%
{%
  name={\ensuremath{\mathbf{o}}},
  description={Observation from simulated characters or robots},
  sort={o}
}

\newglossaryentry{state}%
{%
  name={\ensuremath{\mathbf{s}}},
  description={State of simulated characters or robots},
  sort={s}
}

\newglossaryentry{action}%
{%
  name={\ensuremath{\mathbf{a}}},
  description={Actions applied to characters or robots},
  sort={a}
}

\newglossaryentry{state_space}%
{%
  name={\ensuremath{\mathcal{S}}},
  description={The state space of characters or robots},
  sort={S}
}

\newglossaryentry{observation_space}%
{%
  name={\ensuremath{\mathcal{O}}},
  description={The observation space of characters or robots},
  sort={o}
}

\newglossaryentry{action_space}%
{%
  name={\ensuremath{\mathcal{A}}},
  description={The action space of characters or robots},
  sort={A}
}

\newglossaryentry{reward_function}%
{%
  name={\ensuremath{\mathcal{R}}},
  description={Reward function},
  sort={r}
}

\newglossaryentry{task_reward_function}%
{%
  name={\ensuremath{\mathcal{R}_{task}}},
  description={Task reward},
  sort={r}
}

\newglossaryentry{protect_reward_function}%
{%
  name={\ensuremath{\mathcal{R}_{protect}}},
  description={Safety reward},
  sort={r}
}

\newglossaryentry{transition}%
{%
  name={\ensuremath{\mathcal{T}}},
  description={Transition function of MDP},
  sort={T}
}

\newglossaryentry{observation_function}%
{%
  name={\ensuremath{\mathcal{P}}},
  description={observation function},
  sort={P}
}

\newglossaryentry{init_distribution}%
{%
  name={\ensuremath{p_0}},
  description={Initial state distribution},
  sort={p_0}
}

\newglossaryentry{model_parameters}%
{%
  name={\ensuremath{\boldsymbol{\mu}}},
  description={A vector of physical parameters defining the dynamic model (\eg friction coefficient)},
  sort={mu}
}

\newglossaryentry{curriculum_variable}%
{%
  name={\ensuremath{\vc{x}}},
  description={A vector representing the level of assistance used during the curriculum for learning locomotion. )},
  sort={x}
}

\title{\LARGE \bf
Protective Policy Transfer
}

\author{Wenhao Yu$^{1*}$ C. Karen Liu$^{2}$ and Greg Turk$^{1}$% <-this % stops a space
\thanks{$^{1}$Georgia Institute of Technology, Atlanta, Georgia, USA
        {\tt\small \{wenhaoyu,turk\}@cc.gatech.edu}}%
\thanks{$^{2}$Stanford University,  Stanford, California, USA
        {\tt\small karenliu@cs.stanford.edu}}%
\thanks{$^{*}$Currently at Robotics at Google}%
}

\begin{document}

\maketitle
\thispagestyle{empty}
\pagestyle{empty}

%%%%%%%%%%%%%%%%%%%%%%%%%%%%%%%%%%%%%%%%%%%%%%%%%%%%%%%%%%%%%%%%%%%%%%%%%%%%%%%%
\begin{abstract}

Being able to transfer existing skills to new situations is a key capability when training robots to operate in unpredictable real-world environments. A successful transfer algorithm should not only minimize the number of samples that the robot needs to collect in the new environment, but also prevent the robot from damaging itself or the surrounding environment during the transfer process. In this work, we introduce a policy transfer algorithm for adapting robot motor skills to novel scenarios while minimizing serious failures. Our algorithm trains two control policies in the training environment: a task policy that is optimized to complete the task of interest, and a protective policy that is dedicated to keep the robot from unsafe events (e.g. falling to the ground). To decide which policy to use during execution, we learn a safety estimator model in the training environment that estimates a continuous safety level of the robot. When used with a set of thresholds, the safety estimator becomes a classifier for switching between the protective policy and the task policy. We evaluate our approach on four simulated robot locomotion problems and a 2D navigation problem and show that our method can achieve successful transfer to notably different environments while taking the robot's safety into consideration.

\end{abstract}

\section{Introduction}

When performing motor skills in a novel environment, humans tend to act more conservatively at the beginning to avoid potential dangers and gradually take more agile movements to improve our performance as we get more familiar with the situation. This capability of temporarily shifting the focus between task completion and self protection allows us to transfer our existing skills to scenarios we have not seen before in a protective way and is crucial for us to survive in the ever-changing and unpredictable real-world environment.

%Humans have the ability to identify potentially dangerous situations and temporarily shift the focus from task completion to self protection. For example, an experienced driver can quickly recognize hazardous events like an unexpected car merging from a T-intersection and brake to avoid collision \cite{borowsky2010age}.  To perform such reasoning, one needs to make predictions about ``what could go wrong'' given the current observations, and decide what actions to take to avoid undesired outcomes. Furthermore, we can quickly adapt these knowledge to novel situations we have not seen before, which is essential for humans and animals to survive the ever-changing and unpredictable real environments.

This work aims to develop computational tools that allow a robot to also adapt its motor skills to a new scenario safely. We focus on the practical problem of legged locomotion because legged robots are often prone to falling to the ground and damaging themselves due to the fast movements and intricate balance that is involved. Being able to transfer legged locomotion skills safely to new environments is a crucial step towards deploying robots in real-world applications.

There has been extensive research in transferring control policies from one domain to another \cite{TanRSS18, yu2019learning, peng2020learning, tobin2017domain} as well as incorporating safety in robotic control \cite{achiam2017constrained, tessler2018reward, ames2019control, berkenkamp2017safe}. However, few of them address \emph{both safety and transfer}. Existing methods in safety usually considers learning a policy that satisfies certain safety constraints such as velocity limit and obstacle avoidance. However, the optimized policies usually do not transfer easily to novel situations. On the other hand, current transfer learning algorithms often require additional data collection to adapt the policy to new environments, but do not consider the safety issue during the process of adaptation.

In this work, we introduce a transfer learning algorithm that takes the safety of the robot into consideration. We assume that the robot is first trained in a simulated environment, where it can explore different actions and potentially dangerous states. Our transfer learning algorithm consists of four components. We first train a task policy $\pi_{task}$ to achieve optimal performance for the task. Based on states visited by $\pi_{task}$, we train a protective policy $\pi_{protect}$ that protects the robot from unsafe states such as falling on the ground. To determine which policy to use during execution, we learn a safety estimator model $\psi$ that estimates a continuous safety level of the robot. Training of these models is performed in the source environment. Finally, we develop an adaptation procedure to transfer the learned models from source to target environment with at most two unsafe trials. 

\begin{figure}[t]
  \centering
  \includegraphics[width=0.95\linewidth]{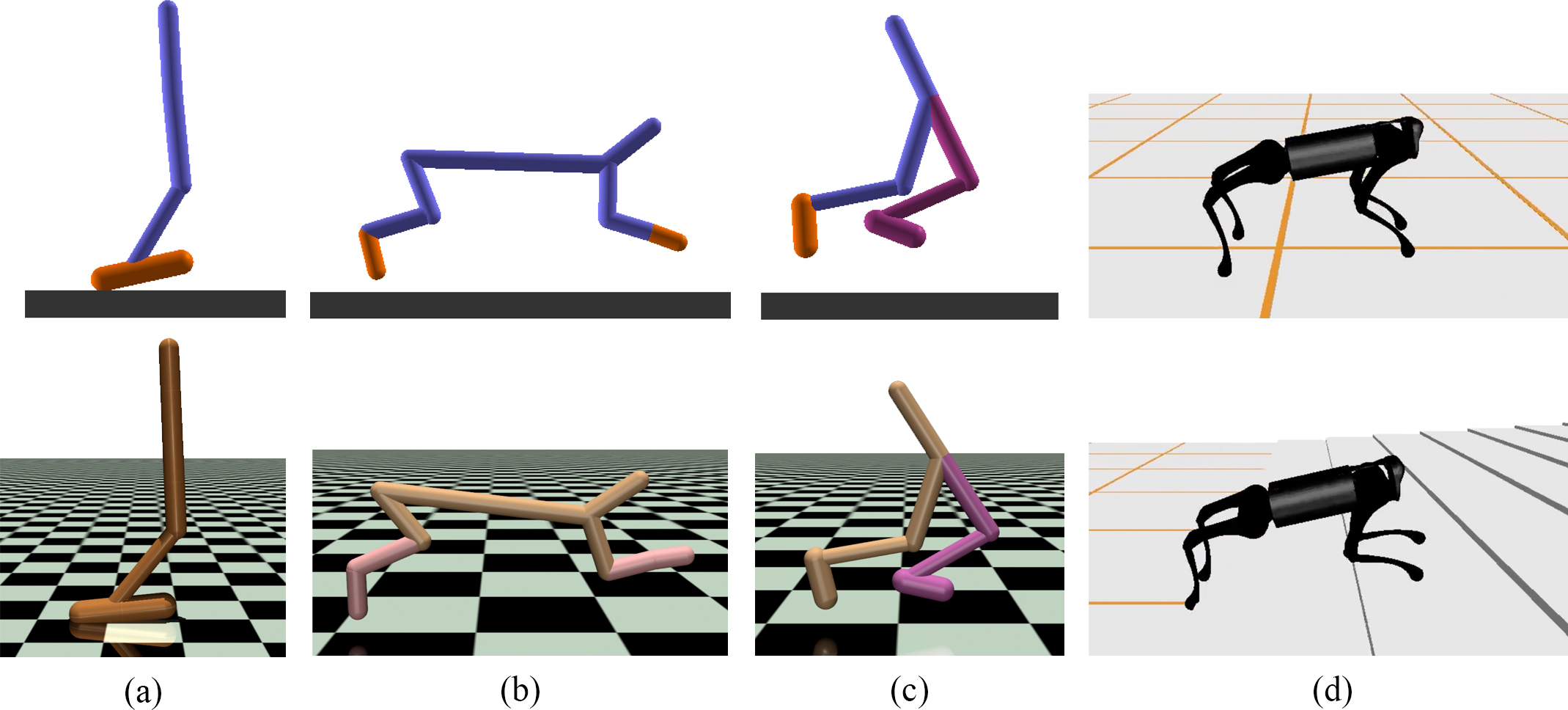}
  \vspace{-3mm}
  \caption{The four tasks used in our experiments: (a) Hopper, (b) HalfCheetah, (c) Walker2D, and (d) 3D Quadruped. Our algorithm trains policies in the source environments (Top row) and transfer them to target environments while minimizing damage to the robot (Bottom row).}
  \label{fig:safe_tasks}
  \vspace{-3mm}
\end{figure}

We evaluate our method of transferring locomotion skills to novel environments for four different robots, as shown in Figure \ref{fig:safe_tasks}. For each robot, we design a variety of testing scenarios inspired by prior work in sim-to-real transfer \cite{TanRSS18}, which creates non-trivial gaps between the training and testing environments of the robot. We demonstrate that our method can successfully complete the task while minimizing the number of failed trials in novels environments, while baseline methods do not achieve successful transfer or require a higher number of failed trials.
\section{Related Work}

Transferring skills learned in a source environment such as computer simulation to novel real-world scenarios is an important direction for creating robots that can be deployed in real-world applications. Recent advances in transfer learning for reinforcement learning have demonstrated some success in deploying simulation-trained policies to real physical robots \cite{TanRSS18, yu2019sim, yu2019learning, peng2020learning, cully2015robots, akkaya2019solving, Siekmann-RSS-20, xie-corl}. One of the key strategies to achieve successful policy transfer is to improve the simulation accuracy by performing system identification \cite{TanRSS18, peng2018sim, akkaya2019solving, Siekmann-RSS-20, xie-corl, hwangbo2019learning}. For example, Tan \etal identified key aspects to the discrepancy between simulated and real robot dynamics such as actuator dynamics and sensor latency \cite{TanRSS18}. By properly measuring these factors and modeling them in the computer simulation, they achieved successful sim-to-real transfer for a real quadruped robot.

Complementary to improving the simulation fidelity, researchers have also developed algorithms to train policies that can work for a large variety of scenarios. One class of these algorithms is to fine-tune the simulation-trained controller with a limited amount of data on the real-robot \cite{yu2019sim, yu2019learning, peng2020learning, Chebotar19, Yang2020Single}. For example, Yu \etal proposed to learn a parameterized controller with a latent space in simulation and optimize the latent input directly on the real hardware \cite{yu2019sim}. Allowing controllers to be fine-tuned using real-world data greatly improves their adaptation performance in novel situations. However, the safety of the robot during real-world experiments is not considered in existing methods. As a result, extra supervision is needed from the experimenter. Our method improved upon the prior methods in this category by considering the safety of the robot during adaptation.

Researchers have also proposed policy optimization methods that takes the safety of the robot into consideration. Some of existing methods enforce the safety of the robot using a model-based approach \cite{garcia2012safe, hans2008safe, perkins2002lyapunov, berkenkamp2017safe, aswani2013provably, ames2019control, cheng2019end}. Given an initial safe but sub-optimal policy, Garcia \etal \cite{garcia2012safe} and Berkenkamp \etal \cite{berkenkamp2017safe} proposed exploration schemes that are provably safe around the current policy and used them to gradually expand the operation region for the policy. Despite being able to provide guarantees on the safety, these methods usually assume the availability of the system dynamics, which may not be true in real-world problems.

Alternatively, researchers have proposed model-free methods for learning policies that satisfy certain safety constraints \cite{achiam2017constrained, tessler2018reward, zhang2020first, schreiter2015safe, berkenkamp2016safe, fan2019safety, eysenbach2017leave}. For example, Achiam \etal introduced a general DRL algorithm, Constrained Policy Optimization (CPO), that solves a constrained MDP problem \cite{achiam2017constrained}. They demonstrated CPO on simulated continuous control tasks with state constraints. Eysenbach \etal proposed to jointly train a policy to complete the task and a policy to reset the robot to the initial state when the robot is in an unsafe state \cite{eysenbach2017leave}. Although these methods can obtain controllers that respects safety constraints in the limit, during the learning process the robot may still be exposed to risky states.

In addition to enforcing general safety constraints, researchers have also investigated more task-dependent safety problems. For example, for humanoid robots, researchers have devised specialized algorithms to reduce the damage they receives during falling \cite{ha2015multiple, fujiwara2002ukemi, fujiwara2006towards, kumar2017learning}. Though effective in handling the falling scenario, these methods may not generalize to other types of failure modes. 
\section{Methods}

%We propose a new method for transferring policies from a source environment to a notably different testing environment while minimizing potential damages to the robot. Our algorithm first trains a task policy $\pi_{task}$ to achieve the desired goal, such as walking forward (Section \ref{ssec:task_pol}). We then train a protective policy $\pi_{protect}$ that is dedicated to keeping the robot safe (Section \ref{ssec:protect_pol}). Using $\pi_{task}$ and $\pi_{protect}$, we train a safety estimation model in the source environment for assessing the safety level of the robot (Section \ref{ssec:ossc}), which is used to select an action from one of the policies during execution. Finally, we develop an adaptation algorithm for tailoring the learned models to the target environment with at most two unsafe trials (Section \ref{ssec:safe_adapt}).

%A key observation we make is that because $\pi_{protect}$ is trained to take the robot to a safer and more balanced pose, it often generalizes to novel situations better than $\pi_{task}$. 

\subsection{Training a task policy}
\label{ssec:task_pol}

We first train a task policy $\pi_{task}$ that performs the desired motor skill using deep reinforcement learning (DRL). We formulate the motor skill learning problem as a Markov Decision Process (MDP), $\mathcal{M}=(\gls*{state_space}, \gls*{action_space}, \gls*{reward_function}, \gls*{transition}, \gls*{init_distribution}, \gamma)$, where $\gls*{state_space}$ is the state space, $\gls*{action_space}$ is the action space, $\gls*{reward_function}: \gls*{state_space} \times \gls*{action_space} \mapsto \mathbb{R}$ is the reward function, $\gls*{transition}: \gls*{state_space} \times \gls*{action_space} \mapsto \gls*{state_space}$ is the transition function, \gls*{init_distribution} is the initial state distribution and $\gamma$ is a discount factor. In DRL, a policy $\pi$ represented by a neural network is searched to maximize the expected return: 
\begin{equation}
    J_{task} = \mathbb{E}_{\tau=(\gls*{state}_0, \gls*{action}_0, \dots, \gls*{state}_T)} \sum_{t=0}^{T} \gamma^t \gls*{reward_function}(\gls*{state}_t, \gls*{action}_t),
\end{equation}
where $\gls*{state}_0 \sim \gls*{init_distribution}$, $\gls*{action}_t \sim \pi(\gls*{state}_t)$ and $\gls*{state}_{t+1}=\gls*{transition}(\gls*{state}_t, \gls*{action}_t)$. In practice, we often only have access to a partial observation $\gls*{observation} \in \gls*{observation_space}$ of the robot's state, where $\gls*{observation_space}$ is the observation space. Thus the policies we train would become $\pi: \gls*{observation_space} \mapsto \gls*{action_space}$. we use Proximal Policy Optimization (PPO) \cite{schulman2017proximal} to optimize the motor skill policies, though our algorithm is agnostic to the specific choice of DRL algorithm. During training of $\pi_{task}$, we randomize the dynamics of the simulated robot, also known as Dynamics Randomization (DR) \cite{peng2018sim}, to improve its robustness. However, as we show in our experiments, using DR alone is not sufficient to achieve successful transfer. 

\subsection{Training a protective policy}
\label{ssec:protect_pol}

% \begin{algorithm}[t]
% \caption{Learning Protective Policy}\label{alg:learn_usp}
% \begin{algorithmic}[1]
% % \State // $\pi_{protect}$ is trained entirely in source environment \;
% \State Randomly initialize the weights for $\pi_{protect}$ \;
% \State Run $\pi_{task}$ for $L$ rollouts and store states in $\mathbf{S}$ \;
% % \State For each state $\mathbf{s}$ in $\mathbf{S}$, randomly perturb $\mathbf{s}$ and add it to $S$ \;
% \For{$i=1:K$}
% \State Initialize $\mathbf{R} \leftarrow \{\}$, $t \leftarrow 0$ \;
% \State Randomly choose initial state $\mathbf{s}_0$ from $\mathbf{S}$ \;
% \While{$t \leq MaxStep$}
% \State $\gls*{observation}_t = \gls*{observation_function}(\gls*{state}_t)$
% \State $\gls*{action}_t = \pi_{protect}(\gls*{observation}_t)$ \;
% \State $\gls*{state}_{t+1} = \gls*{transition}^s (\gls*{state}_t, \vc{a}_t)$ \;
% \State $r$, \texttt{terminate} $= \gls*{protect_reward_function}(\gls*{state}_t, \gls*{action}_t)$
% \State Push $(\gls*{observation}_t, \gls*{action}_t, r)$ into $\mathbf{R}$ \;
% \If{\texttt{terminate}}
% \State Sample $\gls*{state}_{t+1}$ from $\mathbf{S}$  \;
% \EndIf
% \EndWhile
% \State Update $\pi_{protect}$ with $\mathbf{R}$ using DRL \;
% \EndFor
% \Return{$\pi_{protect}$}
% \end{algorithmic}
% \end{algorithm}

To prevent potential damage to the robot, we train a protective policy $\pi_{protect}$ that is dedicated to keeping the robot within a set of safe states $\vc{S}_{safe}$. The boundary of $\vc{S}_{safe}$ can be precisely defined by general inequality constraints on state variables. For example, we can reuse the inequality constraints defined for early termination conditions as the boundary of $\vc{S}_{safe}$. Concretely, in a locomotion problem, a state where the robot falls to the ground will terminate the rollout and is considered outside of $\vc{S}_{safe}$. To train $\pi_{protect}$, we design a reward function: $\gls*{protect_reward_function}(\gls*{state}, \gls*{action}) = w_{alive} r_{alive} + w_{action} r_{action} + w_{task} r_{task}$, where $r_{alive}$ is a constant reward given to the robot, $\gls*{state}_t \in \vc{S}_{safe}$, $r_{action} = ||\gls*{action}||^2_2$ is a regularization term for limiting the magnitude of the actions, and $r_{task} = \min(\gls*{task_reward_function}(\gls*{state}, \gls*{action}), 0)$ discourages the policy from exhibiting behaviors that hurt the desired task. $w_{alive}$, $w_{action}$, $w_{task}$ modulate the relative importance of the reward terms. Note that $r_{task}$ is always negative and only penalizes the agent when it is acting in opposition to the task, e.g. if the robot walks backward when the task is to move forward. We find this helps us train protective policies that are more compatible with the task policies. Section \ref{ssec:setup} describes the weights and the $\vc{S}_{safe}$ for different problems demonstrated in this work.

$\pi_{protect}$ needs to work for all states that the robot might encounter during adaptation. However, we do not know beforehand what states the real robot is going to visit. In our work, we approximate this state distribution by running $\pi_{task}$ with added noise in the source environment to collect a set of states $\mathbf{S}$ that are relevant to the task. We then use $\mathbf{S}$ as the initial state distribution when training $\pi_{protect}$.

An important assumption we make is that $\pi_{protect}$ not only minimizes the potential damage to the robot, but also generalizes to novel environments better than $\pi_{task}$. For the locomotion tasks, this is a reasonable assumption because $\pi_{protect}$ is specialized to prevent the robot from falling over and often learns to take the robot to stable poses such as standing upright, which are usually less sensitive to different dynamics. We validate this assumption in Section \ref{ssec:generalization_compare}. We do note that for other transfer problems, this assumption may not always hold. However, our proposed framework of protective policy transfer is agnostic to how $\pi_{protect}$ is obtained as long as it satisfies the generalization assumption.

\subsection{Training a one-step safety estimator}

\label{ssec:ossc}

\begin{algorithm}[t]
\caption{Learning OSSE}\label{alg:learn_ossc}
\begin{algorithmic}[1]
\State Randomly initialize weights $\psi$ for OSSE \;
\State Obtain the value function $V_{protect}$ of $\pi_{protect}$ \;
\State Run $\pi_{task}$ for $L$ rollouts and store states in $\mathbf{S}$ \;
\State Run $\pi_{protect}$ for $M$ rollouts from states drawn from $\mathbf{S}$ \;
\State Store the generated states $\gls*{state}_t$ to a buffer $\mathbf{B}$ \;
\For{each $\gls*{state}_t$ in $\mathbf{B}$}
\State Run $\pi_{task}$ from $\gls*{state}_t$ and get $(\gls*{observation}_t, \gls*{observation}_{t+1})$
\State Add $\gls*{observation}_t$ to training inputs $T_{input}$\;
\State Add $V_{protect}(\gls*{observation}_{t+1})$ to training labels $T_{label}$\;
\EndFor
\State Optimize $\psi$ with $T_{input}$ and $T_{label}$ \;
\State\Return{$\psi$}
\end{algorithmic}
\end{algorithm}

$\pi_{task}$ outputs actions that optimize the task reward but do not transfer to unseen environments. On the other hand, $\pi_{protect}$ generates actions that keep the robot safe and can generalize to novel scenarios, yet do not accomplish the desired task. Can we combine them to achieve protective policy transfer in the target environment? Naively interpolating the actions from the two policies is likely to result in suboptimal behaviors that are neither functional nor safe. In this work, we propose to combine the task and protective policies by switching between $\pi_{task}$ and $\pi_{protect}$ based on an estimation of how ``safe'' the robot is: we use $\pi_{task}$ if the robot is sufficiently safe, and use $\pi_{protect}$ otherwise. A natural way to perform this estimation is to use the value function $V_{protect}$ from $\pi_{protect}$ as it measures the expected return of the protective policy. However, $V_{protect}$ may not be sufficient for ensuring safety during transfer. In particular, because $V_{protect}$ only considers actions from $\pi_{protect}$ and does not take $\pi_{task}$ into consideration, a state that is considered safe by $V_{protect}$ may not be safe anymore after taking an action from $\pi_{task}$. This will cause over-estimation of the safety for the robot, leading to unsafe behaviors. To address this issue, we train a predictive model that computes the safety estimation for a state assuming that we take one step from $\pi_{task}$ and then follow $\pi_{protect}$. We call this new model the one-step safety estimator (OSSE). A more detailed description of the algorithm can be found in Algorithm \ref{alg:learn_ossc}. 

\subsection{Protective adaptation to testing environment}
\label{ssec:safe_adapt}

The trained OSSE provides a continuous estimation $\psi(\gls{observation})$ for the safety-level of the robot. To make a binary decision for which policy to use, we further threshold the OSSE predictions to produce a binary output. We find that using a single threshold to perform the policy selection often leads to oscillations between the two policies, causing the robot to go to states that are unfamiliar to either policies. To mitigate this issue, we define two thresholds $\kappa_{task}$ and $\kappa_{protect}$ and classify the robot into one of two operation modes: \texttt{protect} and \texttt{task}. The operation mode of the robot is updated by the OSSE prediction and the thresholds: if the robot is in the \texttt{protect} mode and $\psi(\gls*{observation}) > \kappa_{protect}$, it means the robot is in a sufficiently safe state and we switch to \texttt{task} mode. Similarly, if the robot is in the \texttt{task} mode and $\psi(\gls*{observation}) < \kappa_{task}$, then the robot is not safe anymore and should switch to \texttt{protect} mode. The resulting policy as well as the operation model update rule is shown as follows:

\begin{equation}
% \vspace{-2mm}
\label{eq:combine_policy}
    \pi_{combine}(\gls*{observation}, \texttt{m}) = 
    \begin{cases}
    \pi_{protect}(\gls*{observation}),& \text{if }  \texttt{m} = \texttt{protect}\\
    \pi_{task}(\gls*{observation}),& \text{if }  \texttt{m} = \texttt{task},\\
    \end{cases}
    % \vspace{-2mm}
\end{equation}

\begin{equation}
% \vspace{-2mm}
    \texttt{m} = \begin{cases}
    \texttt{protect},& \text{if } \psi(\gls*{observation}) < \kappa_{task} \text{ \ and \ } \texttt{m} = \texttt{task}\\
    \texttt{task},& \text{if } \psi(\gls*{observation}) > \kappa_{protect} \text{ \ and \ } \texttt{m} = \texttt{protect}.
    \end{cases}
    % \vspace{-2mm}
\end{equation}

\begin{algorithm}[t]
\caption{Protective adaptation algorithm}\label{alg:safe_transfer}
\begin{algorithmic}[1]
\State Input: $\pi_{task}$, $\pi_{protect}$, $\psi$, $\Delta$, $\kappa_{min}$ \;
\State Initialize $\kappa_{task} = 1.0$, $\kappa_{protect} = 1.0$, $\kappa_{task}^* = 1.0$, $\kappa_{protect}^* = 1.0$, $R^*=-\infty$ \;
\While{$\kappa_{task} > \kappa_{min}$}
\State Evaluate performance $R$ of $\pi_{combine}$ \;% with $\kappa_{task}$ and $\kappa_{safe}$ \;
\If{$R > R^*$ }
\State $\kappa_{task}^* = \kappa_{task},  R^*=R$\;
\EndIf
\If{robot is unsafe}
\State $\kappa_{task} = \kappa_{task} + \Delta$
\State break\;
\EndIf
\State $\kappa_{task} = \kappa_{task} - \Delta$
\EndWhile
\While{$\kappa_{protect} > \kappa_{task}$}
\State Evaluate performance $R$ of $\pi_{combine}$ \;% with $\kappa_{task}$ and $\kappa_{safe}$  \;
\If{$R > R^*$ }
\State $\kappa_{protect}^* = \kappa_{protect}, R^*=R$\;
\EndIf
\If{robot is unsafe}
\State $\kappa_{protect} = \kappa_{protect} + \Delta$
\State break\;
\EndIf
\State $\kappa_{protect} = \kappa_{protect} - \Delta$
\EndWhile
\State\Return{$\kappa_{task}^*$, $\kappa_{protect}^*$}
\end{algorithmic}
\end{algorithm}
    
Manually specifying the values for $\kappa_{protect}$ and $\kappa_{task}$ can be challenging: for different control problems and robots, the optimal thresholds can be different. One possible way to find the thresholds is to optimize them in the source environment. However, this will likely lead to a policy that exclusively uses the task policy since it maximizes the task reward in the source environment. We propose a protective adaptation algorithm that automatically finds the values for $\kappa_{protect}$ and $\kappa_{task}$ with at most two unsafe trials in the target environment. Our core idea is to start with the most conservative combination (using $\pi_{protect}$ only) and gradually relax the thresholds to allow the task policy to take actions when the robot is safe. Specifically, we first set $\kappa_{protect}=\kappa_{task}=1$ and gradually reduce $\kappa_{task}$ until the robot encounters the first unsafe trial. This step finds the boundary where $\pi_{task}$ stops transfer to the target environment and needs assistance from $\pi_{protect}$. We then perform a second phase of search where we fix the optimized $\kappa_{task}$ and gradually reduce $\kappa_{protect}$ until the resulting policy leads the robot into an unsafe state. After the search, we will return the $\kappa_{task}$ and $\kappa_{protect}$ that achieve the best task performance. Algorithm \ref{alg:safe_transfer} describes our protective adaptation algorithm in greater details.

% \karen{The algorithm is a lot more complex than what you described here. What about those cases with $R >\bar{R}$? I think we need some text to describe them. How do we set $\bar{R}$? Is that another threshold that needs to be tuned? How did you evaluate $R$? From a single trajectory or an average of trajectories?}. \wenhao{I removed $\bar{R}$ at some point in the code so it should be removed from the algorithm. The inclusion of $\bar{R}$ can potentially reduce the number of failure trials, but its value needs to be tuned for different tasks. We use a single trajectory to evaluate R.}

\section{Results}
\label{sec:results}

\subsection{Experimental Setup}
\label{ssec:setup}

We evaluate our algorithm on four challenging locomotion problems with simulated legged robots, as shown in Figure \ref{fig:safe_tasks}. The first three robots: Hopper, HalfCheetah, and Walker2D, are designed based on OpenAI Gym \cite{brockman2016openai}. For the fourth example, we use a 3D quadruped robot, the Unitree A1 (Figure \ref{fig:safe_tasks}(d)). A1 is equipped with $12$ motors: two at each hip joint and one for each knee joint. The motors are controlled by Proportional Derivative (PD) controllers with desired motor angles $\bar{\vc{q}}$ as input. We train locomotion policies that output $\bar{\vc{q}}$ and constrain the change of each dimension in $\bar{\vc{q}}$ between consecutive timesteps to be smaller than $0.2$ for encouraging smooth motion. In all examples, the policies observe the IMU sensor data, motor angles, and motor velocities of the robot.

\begin{table}[h]
\small
  \caption{Task reward functions.}
  \label{table:task_reward_weights}
  \centering
  \resizebox{0.75\columnwidth}{!}{
  \begin{tabular}{lcccc}
    \toprule
    & Hopper & HalfCheetah & Walker2D & A1       \\
    \midrule
    $v_{max}$        & $\infty$        & $\infty$     & $\infty$ & 2.5 \\
    $w_{vel}$        & 1        & 1     & 1     &3\\
    $w_{action}$     & 0.03     & 0.5   & 0.05 & 0.025 \\
    $w_{knee}$       & 0.5      & 0.5   & 0 & 0 \\
    $w_{hip}$        & -        & -     & - & 1 \\
    $w_{dev}$        & -        & -     & - & 5\\
    \bottomrule
  \end{tabular}
  }
  \vspace{-2mm}
\end{table}

%In all three tasks, the observation space is of form: $\vc{o} = [h, \dot{h}, \dot{x}, \theta, \dot{\theta}, \vc{q}, \dot{\vc{q}}]$, where $h$ is the height of the robot, $x$ is the displacement of the root in the forward direction, $\theta$ is the pitch of the robot, and $\vc{q}$ is the joint angles. We also include the time-derivatives of these quantities denoted as $\dot{h}$, $\dot{x}$, $\dot{\theta}$, and $\dot{\vc{q}}$. The robots are controlled through applied torques to the joints. We use the same torque limit as in OpenAI Gym \cite{brockman2016openai} except for the ankle joint of the Walker2D, which we tuned down from $100$ Nm to $20$ Nm. We find this to help learn a better gait for the robot.

%For the fourth example, we test our algorithm on a 3D quadruped robot, the Unitree A1 (Figure \ref{fig:safe_tasks}(d)). A1 is equipped with $12$ motors: two at each hip joint and one at each knee joint. We design the observation space of A1 to be: $\vc{o} = [\phi, \theta, \dot{\phi}, \dot{\theta}, \dot{\psi}, \dot{\vc{c}}, \vc{q}, \dot{\vc{q}}, \bar{\vc{q}}]$, where $\theta$ is the roll of the robot base, $\psi$ is the yaw of the robot and $\vc{c}$ is the CoM position of the robot. The motors are controlled by Proportional Derivative (PD) controllers with desired motor angles denoted as $\bar{\vc{q}}$. We train locomotion policies that outputs the desired motor angles $\bar{\vc{q}}$ and constrain the change of each dimension in $\bar{\vc{q}}$ between consecutive timesteps to be smaller than $0.2$ in order to encourage smoother motion.

We design the following task reward function tor training locomotion skills: $\gls*{task_reward_function}(\gls*{state}, \gls*{action}) = 1 + w_{vel}\min(\dot{x}, v_{max}) - w_{action} ||\gls*{action}||_2^2 - w_{knee} r_{knee} - w_{hip} |q_{hip}| - w_{dev} |y|$, where $v_{max}$ caps the velocity reward and prevents the robot from exploiting unrealistic movements to achieve fast moving speed, $r_{knee}$ is a binary reward denoting whether the knee joint of the robot is at the limit, $q_{hip}$ is the motor angle for the hip abduction and adduction, which discourages the A1 robot to walk with its legs spread or crossed, and $y$ is the deviation in the left-right direction. Table \ref{table:task_reward_weights} lists the rewards parameters used in our experiments. For training the protective policy $\pi_{protect}$ in all tasks, we use $\gls*{protect_reward_function}(\gls*{state}, \gls*{action}) = 4.0 + 0.3 ||\gls*{action}||^2_2 - 0.2  \min(\gls*{task_reward_function}(\gls*{state}, \gls*{action}), 0)$.

%We note that these adjustments on the models and reward functions are done with the default setting to ensure that reasonable policies can be trained and are not optimized for transfer performance.

\begin{table}[h]
\small
  \caption{Domain Randomization Setting.}
  \label{table:domrand}
  \resizebox{\columnwidth}{!}{
  \centering
  \begin{tabular}{lcccc}
    \toprule
    % & \multicolumn{2}{c}{Hopper} & \multicolumn{2}{c}{HalfCheetah} & \multicolumn{2}{c}{Walker2D} & \multicolumn{2}{c}{A1}       \\
    & Hopper & HalfCheetah & Walker2D & A1       \\
    \midrule
    Friction        & [0.2, 1.0]      & [0.4, 1.5]   & [0.2, 1.0] & [0.4, 1.5] \\
    Restitution     & [0.0, 0.3]      & [0.0, 2.0]   & [0.0, 0.8] & [0.0, 0.8] \\
    Mass            & [2.0, 10.0]kg  & [60,  150]\% & [2.0, 10.0]kg & [60,  150]\% \\
    Joint damping   & [0.5, 3.0]      & [70, 130]\% & [0.01, 0.2] & - \\
    Joint Stiffness & -        & [70, 130]\% & - & -\\
    Torque limit    & [100, 300]Nm & [60,150]\% &  - & [15,25]Nm \\
    Position gain   & -      & -     & -     & [60,  150]\% \\
    \bottomrule
  \end{tabular}
  \vspace{-2mm}
  }
\end{table}

\begin{figure*}[t!]
  \centering
  \includegraphics[width=0.9\linewidth]{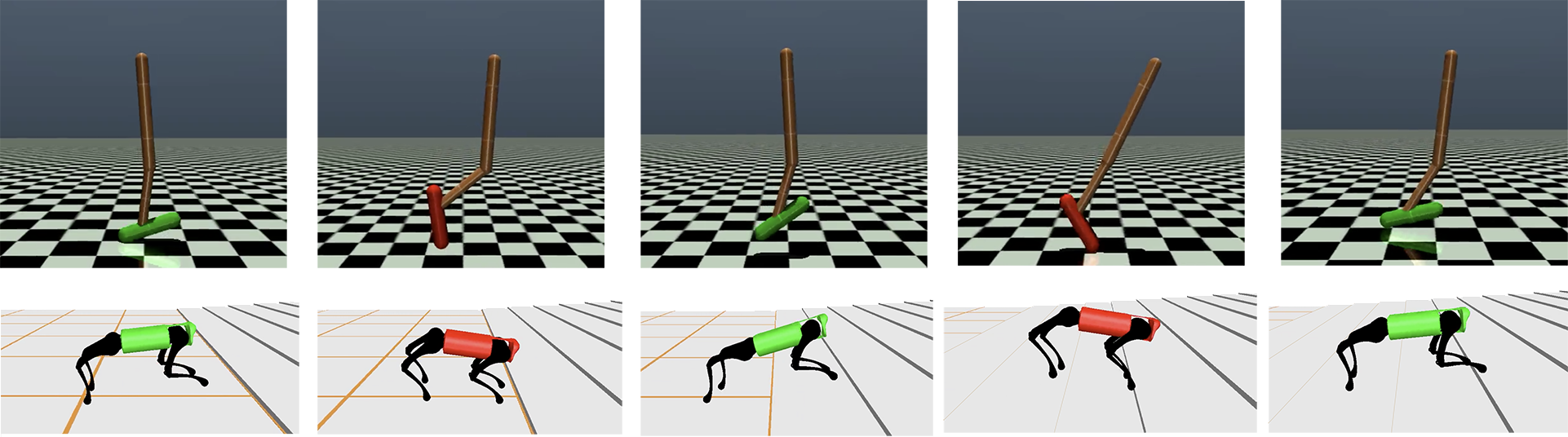}
  \vspace{-3mm}
  \caption{Example sequences of our approach for the Hopper (top) and A1 (bottom) robots. Green bodynode means that $\pi_{task}$ is being used while red bodynodes means $\pi_{protect}$ is being used.}
  \label{fig:example}
  \vspace{-2mm}
\end{figure*}

During training, we perform dynamics randomization \cite{peng2018sim} to improve the robustness of the policies. Table \ref{table:domrand} summarizes the parameters and the range we use for domain randomization for all four tasks. Note that for HalfCheetah and A1 we randomize some parameters using percentage of the default values instead of absolute values. This is to avoid un-physical settings during randomization for these robots.

\begin{figure*}[t]
  \centering
  \includegraphics[width=0.95\linewidth]{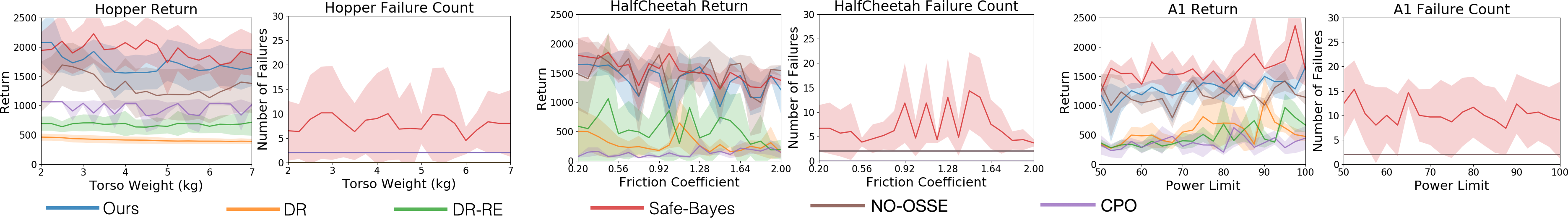}
  \vspace{-3mm}
  \caption{Comparison with baseline methods for Hopper (left), HalfCheetah (middle), and A1 robot (right) in performance and the number of unsafe trials during transfer. We train each policy for five random seeds (one standard deviation shown as shaded area). DR and DR-RE have zero failed trial count since they do not perform adaptation. Our method and NO-OSSE both allow two failed trials.}
  \label{fig:all_results}
  \vspace{-2mm}
\end{figure*}

We define the robot to be safe when it is not falling to the ground or losing balance. Concretely, we specify the safety state set based on three conditions: 1) the robot is allowed to touch the ground only with its feet, 2) the height of the robot should be higher than $h_{min}$, and 3) the roll, pitch, and yaw ($\psi$, $\theta$, $\phi$) of the robot base should not exceed $\psi_{max}, \theta_{max}, \phi_{max}$. For Hopper, we set $h_{min}=0.75$ and $\theta_{max} = 0.8$. For Walker2D, we set $h_{min} = 0.8$ and $\theta_{max} = 1.0$. For HalfCheetah, we set $h_{min} = 0.0$ and $\theta_{max} = 1.3$. And for A1, we have $h_{min} = 0.1$, $\psi_{max} = 0.4$, $\theta_{max}=0.6$, and $\phi_{max}=1.0$. These conditions are also used as the terminal condition during training, which is commonly done in locomotion learning.

We use Proximal Policy Optimization (PPO) \cite{schulman2017proximal} for training both $\pi_{task}$ and $\pi_{protect}$. We represent both policies using neural networks with two fully-connected layers, each with 64 neurons. In our implementation, we use an ensemble of three models for both $V_{protect}$ and OSSE, where each model is represented by a three-layer neural network with $256$, $128$, and $64$ neurons respectively. The final estimated safety level of OSSE is computed as the average of the three models in the ensemble. We use \texttt{tanh} activation for all models. For each method, we run five random seeds and report mean and one standard deviation for the performance. Figure \ref{fig:example} illustrates some of the results by our algorithm for the Hopper and A1 robot in the testing environment. The results can be better seen in the supplementary video.

% we also re-optimize $V_{protect}$ to predict the normalized time-to-failure: $\text{nttf}(\gls*{state}, \pi_{protect}) = \frac{\text{ttf}(\gls*{state}, \pi_{protect})}{H}$, where $\text{ttf}(\cdot)$ is the time-to-failure from the state $\gls*{state}$ using $\pi_{protect}$ and $H$ is the maximum horizon of a rollout ($H=1000$ in our experiments).

\subsection{Transfer tasks}

We design a variety of testing environments for each robot to evaluate the ability of our algorithm to transfer to new situations. For Hopper, Walker2d, and HalfCheetah, we train the policies in one physics simulator, DART \cite{lee2018dart}, and transfer them to a different simulator, MuJoCo \cite{todorov2012mujoco}. As discussed in prior work, the discrepancies between DART and MuJoCo presents significant challenge in transferring policies from one to the other \cite{yu2018policy}. Furthermore, we add latency of $8$ms, $50$ms, and $16$ms to the testing environments of the three robots while no latency is added during training.

For the A1 robot, we build the simulated robot in DART and create two gaps between the training and testing environments. First, we introduce a power limit to each motor in the testing environments. The power for each robot is measured by $\tau \dot{q}$, where $\tau$ is the torque applied by the motor. This effectively creates a speed-dependent torque limit. As shown in prior work \cite{TanRSS18, hwangbo2019learning}, mismatch in the actuator modeling is one of the major sources of the "Reality Gap". Second, the terrain of the testing environment is modified to be a sequence of steps (0.5m $\times$ 0.03m) going upward, while the training environment uses a flat ground.

To better evaluate the transfer performance of different algorithms, we further vary the physical properties during the evaluation in addition to the gaps introduced above. For the Hopper task, we vary the torso mass between $[2.0, 7.0]$ kg. For the HalfCheetah task in MuJoCo, we vary the restitution coefficient of the ground in $[0.5, 1.0]$. For the Walker2D task, we vary the mass of one foot in $[2.0, 15.0]$ kg. And for the A1 robot, we vary the power limit in $[50, 100]$.

\begin{table}[hb]
\small
\vspace{-2mm}
  \caption{Performance of different methods for all examples. Numbers in parenthesis is the average number of failed trials during adaptation.}
  \label{table:performance}
  \centering
  \resizebox{\columnwidth}{!}{
  \begin{tabular}{lcccccc}
    \toprule
                    & Ours     & DR & DR-RE & NO-OSSE & CPO & Safe-Bayes \\
    \midrule
    Hopper          & 1700.5   &   402.0   &  681.1  & 1081.4 & 966.4 & 2092.8 (7)  \\
    HalfCheetah     & 1402.2 &    290.2  &  538.1 & 1389.6 & 193.2 & 1536.9 (9.2)  \\
    Walker2D        & 1656.4 & 1000.4 & 1339.1 & 1383.4 & 626.4 & 1947 (9.3)  \\
    A1              & 1281.6 & 475.18 & 550.9 & 1245.6 & 361.3 & 1615.75 (10.3) \\
    \bottomrule
  \end{tabular}
  }
\end{table}

\subsection{Baseline methods}

We compare our method to four baseline methods. 

1) \textit{Dynamics Randomization (DR)}: a commonly used sim-to-real technique where a robust policy is trained with randomized dynamics.

2) \textit{Dynamics Randomization with Reward Engineering (DR-RE)}: we increase the alive bonus reward by four times to encourage more robust behaviors. 

3) \textit{Safe-Bayes}: we train $\pi_{task}$, $\pi_{protect}$, and OSSE the same way as in our algorithm and apply Bayesian Optimization to search for the thresholds $\kappa_{safe}$ and $\kappa_{task}$ instead of using our protective adaptation algorithm. 

4) \textit{Constrained Policy Optimization (CPO)}: a safe reinforcement learning algorithm by Achiam \etal, which solves a task policy that satisfies the defined safety constraints \cite{achiam2017constrained}. 

We further ablate our proposed method by using the $V_{protect}$ instead of OSSE for estimating the safety level of the robot, which is denoted as NO-OSSE.

\subsection{Does our method outperform baseline methods?}

\begin{figure}[t]
  \centering
  \includegraphics[width=0.95\linewidth]{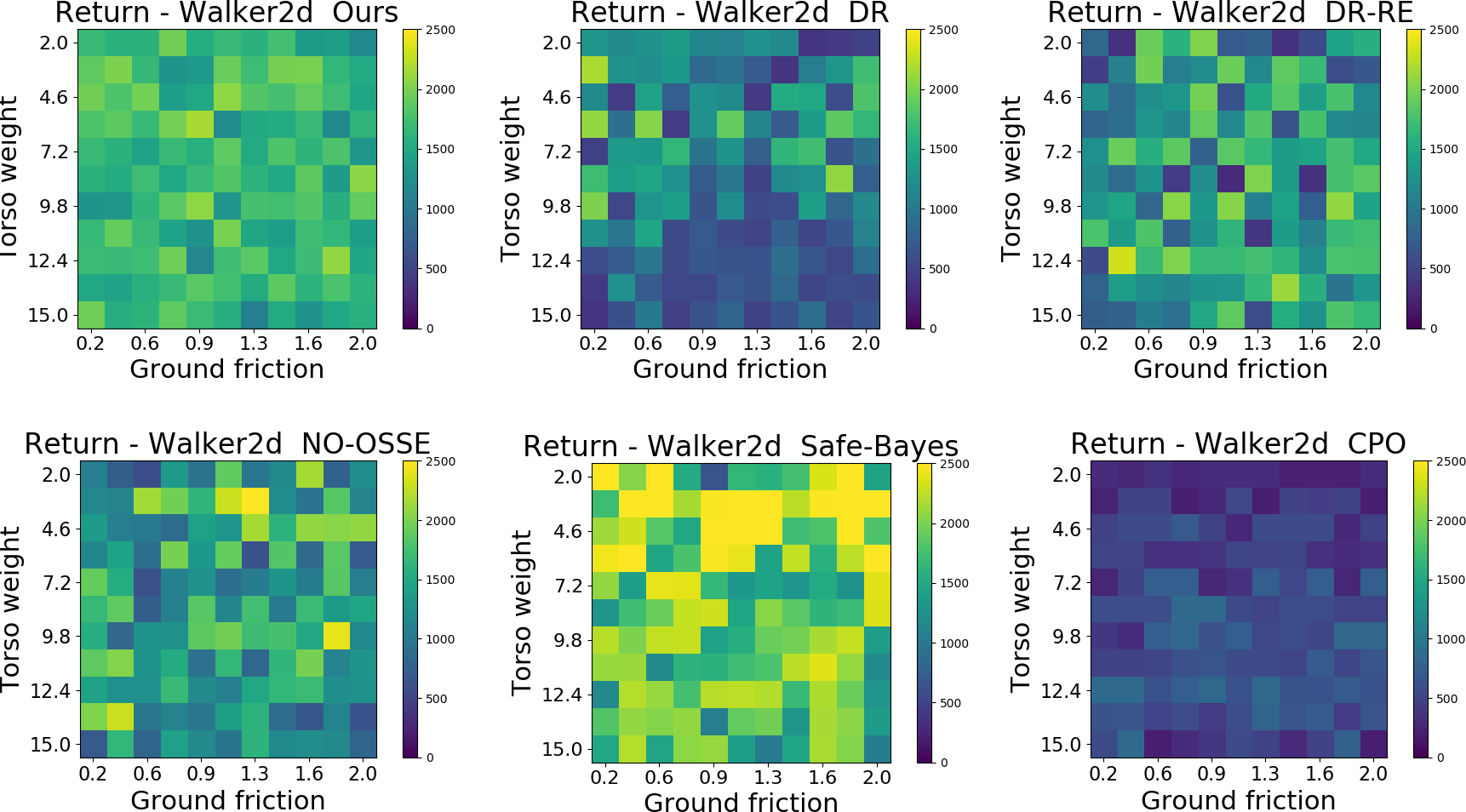}
  \vspace{-4mm}
  \caption{Comparison of our method and the baseline methods for the Walker2D example.}
  \label{fig:walker_result}
  \vspace{-1mm}
\end{figure}

We report the mean task performance of different algorithms in the target environments, as summarized in Table \ref{table:performance}. Since the policies are tested on different dynamics parameters during testing, we further plot the policy performance with respect to the varied parameters in Figure \ref{fig:all_results} and Figure \ref{fig:walker_result}. As shown in these results, our proposed method achieves notably better task performance than DR, DR-RE, and CPO. This shows the importance of adapting the policy in a novel testing environment. We note that it is possible to fine-tune these policies in the testing environment directly using DRL algorithm. However, we did not perform this test because 1) the fine-tuning process will likely introduce many unsafe trials and 2) prior work demonstrated that fine-tuning a DR policy is not very effective when the environment gap is large \cite{yu2018policy}. Meanwhile, applying Bayesian Optimization to find $\kappa_{task}$ and $\kappa_{safe}$ (Safe-Bayes) can in general achieve better task performance than our algorithm. This suggests that the shifting between task performance and robot protection is not simply monotonic as our algorithm hypothesize. However, as shown in Figure \ref{fig:all_results} and Table \ref{table:performance}, Safe-Bayes results in a higher number of failed trials during adaptation. In our experiments, we allow at most $20$ trials during Bayesian Optimization in Safe-Bayes. We have also examined Safe-Bayes using different sample allowance and find that our method achieves better performance when the same budget of unsafe trials is allowed. 

In addition, when the OSSE model is replaced with $V_{protect}$ (NO-OSSE), it results in worse performance for Hopper and Walker2D, but comparable results for HalfCheetah and A1. We hypothesize that this is because when the robot is less stable, it will rely more heavily on taking actions from $\pi_{protect}$ to remain safe. As a result, a state that has high score from $V_{protect}$ may not be safe when we take an action from $\pi_{task}$, leading to lower task reward. On the other hand, for robots that are structurally more stable, $V_{protect}$ can provide a more reasonable estimation of the robot's safety, leading to similar performance between using OSSE and $V_{protect}$.

\begin{figure}[t]
  \centering
  \vspace{-1mm}
  \includegraphics[width=0.85\linewidth]{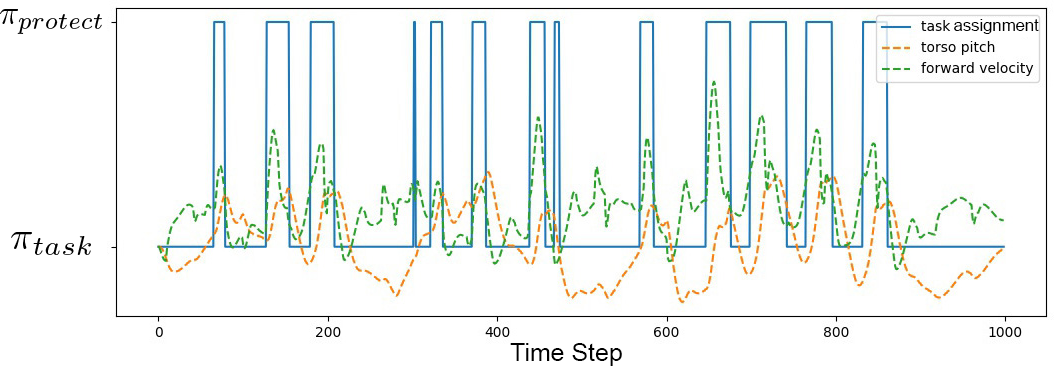}
  \vspace{-2mm}
  \caption{Visualization of one rollout of Hopper task where the blue solid curve represents the policy selection and the dashed curves are the scaled torso pitch and forward velocity of the robot.}
  \label{fig:safe_task_scheduling}
  \vspace{-4mm}
\end{figure}

\subsection{Does our transfer scheme make reasonable policy assignment?}

In this section, we examine the policy assignment strategy that emerges from our method. We apply our method on the MuJoCo Hopper with torso weight of 2.5 kg and record the observations and the policy selection. We then train a logistic regression model to predict the policy selection from the observations. The logistic model shows that torso pitch and forward velocity of the Hopper are the two most predictive features for the policy assignment (Figure \ref{fig:safe_task_scheduling} (a)). For better visualization, we scale down the values for torso pitch and forward velocity. Our policy tends to choose $\pi_{protect}$ when the forward velocity is high and is tilting forward. This makes sense because the combination of high forward velocity and forward tilting indicates that the robot might fall forward. 

Figure \ref{fig:example} shows some examples of our trained policies transferred to the testing environments. The resulting scheduling can be better seen in the supplementary video.

\subsection{Generalization Comparison between $\pi_{task}$ and $\pi_{protect}$}
\label{ssec:generalization_compare}

\begin{table}[t]
\small
  \caption{Generalization comparison for task and protective policy.}
  \label{table:gen_compare}
  \resizebox{\columnwidth}{!}{
  \centering
  \begin{tabular}{lcccccc}
    \toprule
    & \multicolumn{3}{c}{Normalized rollout length} & \multicolumn{3}{c}{Return transfer ratio}       \\
    \cmidrule(r){2-4}  \cmidrule(r){5-7}
    & Hopper     & Cheetah & Walker & Hopper     & Cheetah & Walker  \\
    \midrule
    $\pi_{task}$ & 0.15  & 0.3 & 0.17  &  0.29 &0.14 & 0.39  \\
    $\pi_{protect}$     & 0.75 & 1.0 & 0.97 & 0.99 & 0.94  & 0.95   \\
    \bottomrule
  \end{tabular}
  }
  \vspace{-4mm}
\end{table}

A key assumption we make in our method is that $\pi_{protect}$ can generalize to new scenarios better than $\pi_{task}$. To validate this, we use two metrics to compare $\pi_{protect}$ and $\pi_{task}$. The first one is how long the policy keeps the robot safe. To compute this metric, we run both policies for $200$ rollouts and compute the average rollout lengths in the testing environments. We normalize the lengths using the maximal rollout length (1000). The results can be found in the left of Table \ref{table:gen_compare}. The second metric we use is the ratio of the policy return between testing and training environments, where the return corresponds to the training reward of the policies. We report the average of return ratio over $200$ rollouts. The results is shown in the right half of Table \ref{table:gen_compare}. We sample source and target environment from the same variations described in Section \ref{ssec:setup} when measuring both metrics. We can see that $\pi_{protect}$ achieves better performance in both metrics. This is likely because $\pi_{protect}$ is trained to specialize in staying within safe states. As a result, $\pi_{protect}$ learns to take the robot to stable states such as standing upright, which is less sensitive to different dynamics.

\vspace{-0.1in}

\section{Discussion and Conclusion}

We have presented a transfer learning algorithm for protectively adapting a control policy to novel scenarios. Our algorithm trains three models in the source environment: a task policy that optimizes the task reward, a protective policy that protects the robot form unsafe states, and a safety estimation model for assessing the safety-level of the robot. To transfer these models to novel testing scenarios, we introduce a protective adaptation algorithm that minimizes the unsafe trials during adaptation. We demonstrate that our algorithm can overcome large modeling errors for four different simulated transfer problems and achieve better performance than baseline methods in safely adapting the policy to new environments.

Our algorithm achieves protective policy transfer at the cost of exhibiting conservative behaviors with suboptimal task performance. An important future direction is thus to improve the task performance of our algorithm while keeping the unsafe trials low during transfer. Another interesting future direction is to extend our approach to more diverse control problems such as navigation or manipulation.

%%%%%%%%%%%%%%%%%%%%%%%%%%%%%%%%%%%%%%%%%%%%%%%%%%%%%%%%%%%%%%%%%%%%%%%%%%%%%%%%

\addtolength{\textheight}{-1cm}   % This command serves to balance the column lengths

%%%%%%%%%%%%%%%%%%%%%%%%%%%%%%%%%%%%%%%%%%%%%%%%%%%%%%%%%%%%%%%%%%%%%%%%%%%%%%%%

%%%%%%%%%%%%%%%%%%%%%%%%%%%%%%%%%%%%%%%%%%%%%%%%%%%%%%%%%%%%%%%%%%%%%%%%%%%%%%%%

%%%%%%%%%%%%%%%%%%%%%%%%%%%%%%%%%%%%%%%%%%%%%%%%%%%%%%%%%%%%%%%%%%%%%%%%%%%%%%%%

% \section*{ACKNOWLEDGMENT}

%%%%%%%%%%%%%%%%%%%%%%%%%%%%%%%%%%%%%%%%%%%%%%%%%%%%%%%%%%%%%%%%%%%%%%%%%%%%%%%%

\bibliographystyle{IEEEtran}
\bibliography{reference}

\end{document}